\documentclass[10pt,twocolumn,letterpaper]{article}

\usepackage{cvpr}
\usepackage{times}
\usepackage{epsfig}
\usepackage{graphicx}
\usepackage{amsmath}
\usepackage{amssymb}
\usepackage{mathrsfs}

\usepackage[pagebackref=true,breaklinks=true,letterpaper=true,colorlinks,bookmarks=false]{hyperref}

 \cvprfinalcopy 

\def\cvprPaperID{1944} 

\ifcvprfinal\fi
\begin{document}

\title{Deep Reinforcement Learning of Volume-guided Progressive View Inpainting \\for 3D Point Scene Completion from a Single Depth Image}

\author{$ ^{1,3} $Xiaoguang Han, $ ^{2,3} $Zhaoxuan Zhang, $ ^{3,4} $Dong Du, $ ^{1,3} $Mingdai Yang, $ ^{5} $Jingming Yu, $ ^{5} $Pan Pan\\
	$ ^{2} $Xin Yang, $ ^{4} $Ligang Liu, $ ^{6} $Zixiang Xiong, $ ^{1,3} $Shuguang Cui\\
	\\
$ ^{1} $The Chinese University of Hong Kong(Shenzhen), $ ^{2} $Dalian University of Technology\\
$ ^{3} $Shenzhen Research Institute of Big Data, $ ^{4} $University of Science and Technology of China\\
$ ^{5} $Alibaba Group, $ ^{6} $Texas A\&M University
}

\maketitle

\begin{abstract}
We present a deep reinforcement learning method of progressive view inpainting for 3D point scene completion under volume guidance, achieving high-quality scene reconstruction from only a single depth image with severe occlusion. Our approach is end-to-end, consisting of three modules: 3D scene volume reconstruction, 2D depth map inpainting, and multi-view selection for completion. Given a single depth image, our method first goes through the 3D volume branch to obtain a volumetric scene reconstruction as a guide to the next view inpainting step, which attempts to make up the missing information; the third step involves projecting the volume under the same view of the input, concatenating them to complete the current view depth, and integrating all depth into the point cloud. Since the occluded areas are unavailable, we resort to a deep Q-Network to glance around and pick the next best view for large hole completion progressively until a scene is adequately reconstructed while guaranteeing validity. All steps are learned jointly to achieve robust and consistent results. We perform qualitative and quantitative evaluations with extensive experiments on the SUNCG data, obtaining better results than the state of the art.
\end{abstract}

\begin{figure}
	\centering
	\includegraphics[width=0.45\textwidth]{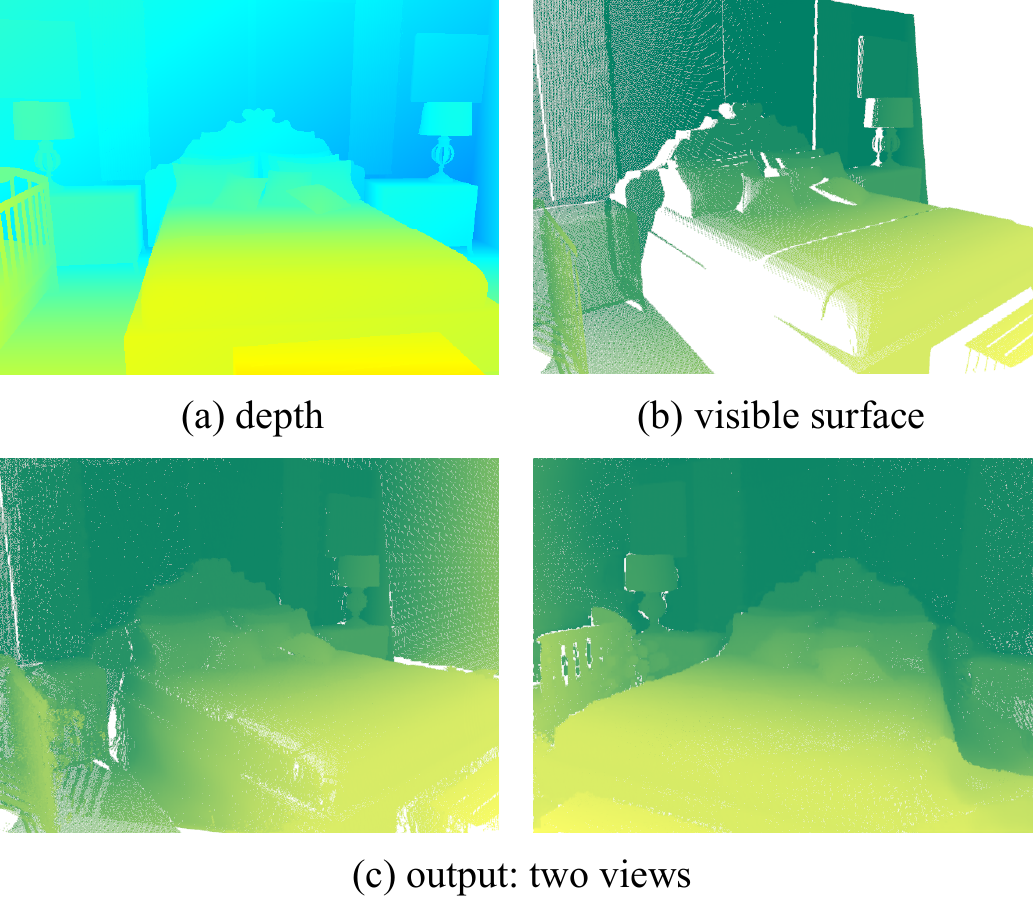}
	\caption{Surface-based scene Completion. (a) A single-view depth map as input; (b) Visible surface from the depth map, which is represented as the point cloud. In our paper, the color of depth and point cloud is for visualization only; (c) Our scene completion results: directly recovering the missing points of the occluded regions. Here we choose two views for a better display.}
	\label{fig:scene_completion}
\end{figure}

\section{Introduction}
Recovering missing information in occluded regions of a 3D scene from a single depth image is a very active research area of late ~\cite{song2017semantic,zhang2018semantic,guo2018view,liu2018see,garbade2018two,wang2018adversarial}. This is due to its importance in robotics and vision tasks such as indoor navigation, surveillance, and augmented reality. Although this problem is mild in human vision system, it becomes severe in machine vision because of the sheer imbalance between input and output information.
One class of popular approaches ~\cite{Shao:2012:IAS:2366145.2366155, Chen:2014:ASM:2661229.2661239, GuptaAGM15, guo2015predicting} to this problem is based on classify-and-search: pixels of the depth map are classified into several semantic object regions, which are mapped to most similar 3D ones in a prepared dataset to construct a fully 3D scene. Owing to the limited capacity of the database, results from classify-and-search are often far away from the ground truth. By transforming the depth map into an incomplete point cloud, Song et al. ~\cite{song2017semantic} recently presented the first end-to-end deep network to map it to a fully voxelized scene, while simultaneously outputting the class labels each voxel belongs to. The availability of volumetric representations makes it possible to leverage 3D convolutional neural networks (3DCNN) to effectively capture the global contextual information, however, starting with an incomplete point cloud results in loss of input information and consequently low-resolution outputs. Several recent works ~\cite{liu2018see,guo2018view,garbade2018two,wang2018adversarial} attempt to compensate the lost information by extracting features from the 2D input domain in parallel and feeding them to the 3DCNN stream. To our best knowledge, no work has been done on addressing the second issue of improving output quality.

Taking an incomplete depth map as input, in this work we advocate the approach of straightforwardly reconstructing 3D points to fill missing region and achieve high-resolution completion (Figure \ref{fig:scene_completion}). To this end, we propose to carry out completion on multi-view depth maps in an iterative fashion until all holes are filled, with each iteration focusing on one viewpoint. At each iteration/viewpoint, we render a depth image relative to the current view and fill the produced holes using 2D inpainting. The recovered pixels are re-projected to 3D points and used for the next iteration. Our approach has two issues: First, different choices of sequences of viewpoints strongly affect the quality of final results because given a partial point cloud, different visible contexts captured from myriad perspectives present various levels of difficulties in the completion task, producing diverse prediction accuracies; moreover, selecting a larger number of views for the sake of easier inpainting to fill smaller holes in each iteration will lead to error accumulation in the end. Thus we need a policy to determine the next best view as well as the appropriate number of selected viewpoints. Second, although existing deep learning based approaches ~\cite{pathak2016context,iizuka2017globally,liu2018image} show excellent performance for image completion, directly applying them to depth maps across different viewpoints usually yields inaccurate and inconsistent reconstructions. The reason is because of lack of global context understanding.
To address the first issue, we employ a reinforcement learning optimization strategy for view path planning. In particular, the current state is defined as the updated point cloud after the previous iteration and the action space is spanned by a set of pre-sampled viewpoints chosen to maximize 3D content recovery. The policy that maps the current state to the next action is approximated by a multi-view convolutional neural network (MVCNN) ~\cite{su15mvcnn} for classification. The second issue is handled by a volume-guided view completion deepnet. It combines one 2D inpainting network ~\cite{liu2018image} and another 3D completion network ~\cite{song2017semantic} to form a joint learning machine. In it low-resolution volumetric results of the 3D net are projected and concatenated to inputs of the 2D net, lending better global context information to depth map inpainting. At the same time, losses from the 2D net are back-propagated to the 3D stream to benefit its optimization and further help improve the quality of 2D outputs. As demonstrated in our experimental results, the proposed joint learning machine significantly outperforms existing methods quantitatively and qualitatively.

In summary, our contributions are
\vspace{-.08in}
\begin{itemize}
\itemsep -.05in
  \item The first surface-based algorithm for 3D scene completion from a single depth image by directly generating the missing points.
  \item A novel deep reinforcement learning strategy for determining the optimal sequence of viewpoints for progressive scene completion.
  \item A volume-guided view inpainting network that not only produces high-resolution outputs but also makes full use of the global context.
\end{itemize}

\begin{figure*}
	\centering
	\includegraphics[width=0.975\textwidth]{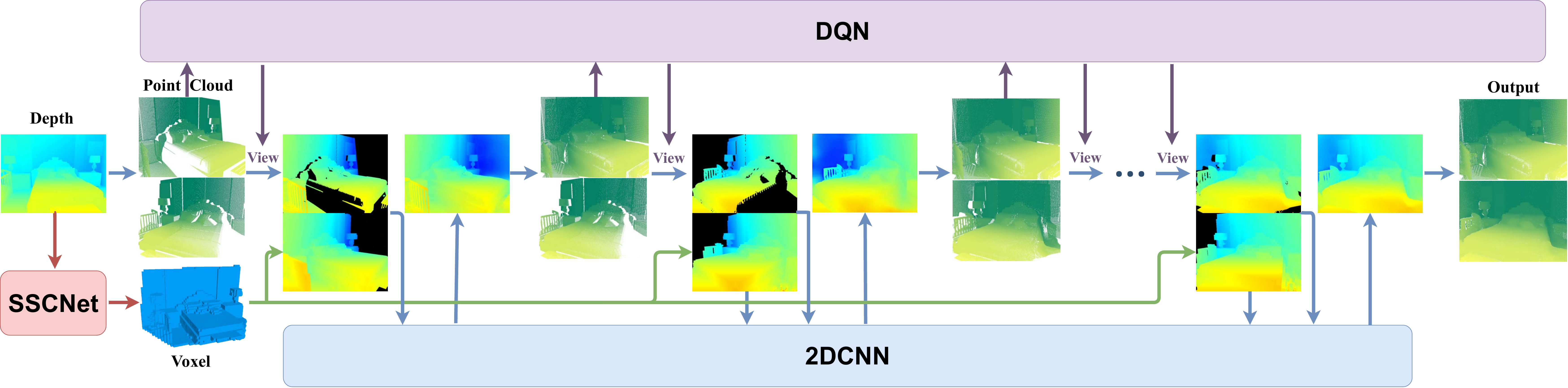}
	\caption{The pipeline of our method. Given a single depth image $D_0$, we convert it to a point cloud $P$, here shown in two different views. DQN is used to seek the next-best-view, under which the point cloud is projected to a new depth image $D_1$, causing holes. In parallel, the $P$ is also completed in volumetric space by SSCNet, resulting in $V$. Under the view of $D_1$, $V$ is projected and guide the inpainting of $D_1$ with a 2DCNN network. Repeating this process several times, we can achieve the final high-quality scene completion.}
	\label{fig:pipeline}
\end{figure*}

\section{Related Works}
Many prior works are related to scene completion. The literature review is conducted in the following aspects.

\textbf{Geometry Completion}
Geometry completion has a long history in 3D processing, known for cleaning up broken single objects or incomplete scenes. Small holes can be filled by primitives fitting\cite{schnabel2009completion, li2011globfit}, smoothness minimization\cite{sorkine2004least, zhao2007robust, kazhdan2013screened}, or structures analysis\cite{mitra2006partial, sipiran2014approximate, sung2015data}. These methods however seriously depend on prior knowledge. Template or part based approaches can successfully recover the underlying structures of a partial input by retrieving the most similar shape from a database, matching with the input, deforming disparate parts and assembling them\cite{shen2012structure, kim2013learning, rock2015completing, sung2015data}. However, these methods require manually segmented data, and tend to fail when the input does not match well with the template due to the limited capacity of the database. Recently, deep learning based methods have gained much attentions for shape completion\cite{rock2015completing, thanh2016field, sharma2016vconv, varley2017shape, dai2017shape, han2017high}, while scene completion from sparse observed views remains challenging due to large-scale data loss in occluded regions. Song et al.\cite{song2017semantic} first propose an end-to-end network based on 3DCNNs, named SSCNet, which takes a single depth image as input and simultaneously outputs occupancy and semantic labels for all voxels in the camera view frustum. ScanComplete\cite{dai2018scancomplete} extends it to handle larger scenes with varying spatial extent. Wang et al.\cite{wang2018adversarial} combine it with an adversarial mechanism to make the results more plausible. Zhang et al.\cite{zhang2018semantic} apply a dense CRF model followed with SSCNet to further increase the accuracy. In order to exploit the information of input images, Garbade et al.\cite{garbade2018two} adopt a two stream neural network, leveraging both depth information and semantic context features extracted from the RGB images. Guo et al.\cite{guo2018view} present a view-volume CNN which extracts detailed geometric features from the 2D depth image and projects them into a 3D volume to assist completed scene inference. However, all these works based on the volumetric representation result in low-resolution outputs. In this paper, we directly predict point cloud to achieve high-resolution completion by conducting inpainting on multi-view depth images.

\textbf{Depth Inpainting}
Similar to geometry completion, researchers have employed various priors or optimized models to complete a depth image\cite{herrera2013depth, liu2012guided, muddala2014depth, thabet20143d, chen2014improved, liu2016building, xue2017depth, zhang2018probability}. The patch-based image synthesis idea is also applied\cite{doria2012filling, gautier2011depth}. Recently, significant progresses have been achieved in image inpainting field with deep convolutional networks and generative adversarial networks (GANs) for regular or free-form holes\cite{iizuka2017globally, liu2018image, yu2018free}. Zhang et al.\cite{zhang2018deep} imitate them with a deep end-to-end model for depth inpainting. Compared with inpainting task on colorful images, recovering missing information from a single depth map is more challenging due to the absence of strong context features in depth maps. To address it, an additional 3D global context is provided in our paper, guiding the inpainting on diverse views to reach more accurate and consistent output.

\textbf{View Path Planing}
Projecting a scene or an object to the image plane will severely cause information loss because of self-occusions. A straightforward solution is utilizing dense views for making up\cite{su15mvcnn, qi2016volumetric, tatarchenko2016multi}, yet it will lead to heavy computation cost. Choy et al.\cite{choy20163d} propose a 3D recurrent neural networks to integrate information from multi-views which decreases the number of views to five or less. Even so, how many views are sufficient for completion and which views are better to provide the most informative features, are still open questions. Optimal view path planning, as the problem to predict next best view from current state, has been studied in recent years. It plays critical roles for scene reconstruction as well as environment navigation in autonomous robotics system\cite{low2006adaptive, blaer2007data, zhou2013dense, wu2014quality}. Most recently, this problem is also explored in the area of object-level shape reconstruction\cite{yang2018active}. A learning framework is designed in \cite{xu20163d}, by exploiting the spatial and temporal structure of the sequential observations, to predict a view sequence for groundtruth fitting. Our work explores the approaches of view path planning for scene completion. We propose to train a Deep Q-Network (DQN)\cite{mnih2015human} to choose the best view sequence in a reinforcement learning framework.

\section{Algorithm}
\noindent \textbf{Overview}

Taking a depth image $D_0$ as input, we first convert it to a point cloud $P_0$, which suffers from severe data loss. Our goal is to generate 3D points to complete $P_0$. The main thrust of our proposed algorithm is to represent the incomplete point cloud as multi-view depth maps and perform 2D inpainting tasks on them. To take full advantage of the context information, we execute these inpainting operations view by view in an accumulative way, with inferred points for the current viewpoint kept and used to help inpainting of the next viewpoint. Assume $D_0$ is rendered from $P_0$ under viewpoint $v_0$, we start our completion procedure with a new view $v_1$ and render $P_0$ under $v_1$ to obtain a new depth map $D_1$, which potentially has many holes. We fill these holes in $D_1$ with 2D inpainting, turning $D_1$ to $\hat{D_1}$. The inferred depth pixels in $\hat{D_1}$ are then converted to 3D points and aggregated with $P_0$ to output a denser point cloud $P_1$. This procedure is repeated for a sequence of new viewpoints $v_2,v_3,...,v_n$, yielding point clouds $P_2,P_3,...,P_n$, with $P_n$ being our final output. Figure \ref{fig:pipeline} depicts the overall pipeline of our proposed algorithm.
Since $P_n$ depends on the view path $v_2,v_3,...,v_n$, we describe in section ~\ref{sec:dqn} a deep reinforcement learning framework to seek the best view path. Before that, we introduce our solution to another critical problem of 2D inpainting, i.e., transforming $D_i$ to $\hat{D_i}$, in section ~\ref{sec:inpaint} first.

\subsection{Volume-guided View Inpainting}
\label{sec:inpaint}

Deep Convolutional Neural Network (CNN) has been widely utilized to effectively extract context features for image inpainting tasks, achieving excellent performance. Although it can be directly applied to each viewpoint independently, this simplistic approach will lead to inconsistencies across views because of lack of global context understandings. We propose a volume-guided view inpainting framework by first conducting completion in the voxel space, converting $P_0$'s volumetric occupancy grid $V$ to its completed version $V^c$. Denote the projected depth map from $V^c$ to the view $v_i$ as $D^c_i$. Our inpainting of the $i_{th}$ view takes both $D_i$ and $D^c_i$ as input and outputs $\hat{D_i}$. As shown in Figure \ref{fig:pipeline}, this is implemented using a three-module neural network architecture consisting of a volume completion network, a depth inpainting network, and a differentiate projection layer connecting them. The details of each module and our training strategy are described below.

\noindent \textbf{Volume Completion} We employ SSCNet proposed in ~\cite{song2017semantic} to map $V$ to $V^c$ for volume completion. SSCNet predicts not only volumetric occupancy but also the semantic labels for each voxel. Such a multi-task learning scheme helps us better capture object-aware context features and contributes to higher accuracy. The readers are referred to ~\cite{song2017semantic} for details on how to set up this network architecture. We train the network as a voxel-wise binary classification task and take the output 3D probability map as $V^c$. The resolution of input is $240\times144\times240$, and the output is $60\times36\times60$.

\noindent\textbf{Depth Inpainting} In our work, the depth map is rendered as a $512\times 512$ grayscale image. Among various existing approaches, the method of ~\cite{liu2018image} is chosen to handle our case with holes of irregular shapes. Specifically, $D_i$ and $D^c_i$ are first concatenated to form a map with $2$ channels. The resulting map is then fed into a U-Net structure implemented with a masked and re-normalized convolution operation (also called partial convolution), followed by an automatic mask-updating step. The output is also in $512\times 512$. We refer the readers to ~\cite{liu2018image} for details of the architecture settings and the design of loss functions.

\noindent\textbf{Projection Layer} As validated in our experiments described in ~\ref{sec:ablation}, the projection of $V^c$ greatly benefits inpainting of 2D depth maps.  We further exploit the benefit of 2D inpainting to volume completion by propagating the 2D loss back to optimize the parameters of 3D CNNs. Doing so requires a differentiable projection layer, which was recently proposed in ~\cite{tulsiani2017ray}. Thus, we connect $V^c$ and $D^c_i$ using this layer. For the sake of notational convenience, we use $V$ to represent $V^c$ and $D$ to represent $D^c_i$. Specifically, for each pixel $x$ in $D$, we launch a ray that starts from the viewpoint $v_i$, passes through $x$, and intersects a sequence of voxels in $V$, noted as $l_{1}, l_{2},..., l_{N_{x}}$. We denote the value of the $k_{th}$ voxel in $V$ as $V_k$, which represents the probability of this voxel being empty. Then, we define the depth value of this pixel $x$ as
\begin{equation}\label{key}
D(x)=\sum^{N_{x}}_{k=1}P^{x}_{k} d_{k}
\end{equation}
where $d_{k}$ is the distance from the viewpoint to voxel $l_{k}$ and $P^{x}_{k}$ the probability of the ray corresponding to $x$ first meets the $l_{k}$ voxel
\begin{equation}\label{key}
P^{x}_{k}=(1-V_{k})\prod^{k-1}_{j=1}V_{j},  \ k=1,2,...,N_{x} 
\end{equation}
The derivative of $ D(x)$ with respect to $V_{k}$ can be calculated as
\begin{equation}\label{key}
\frac{\partial D(x)}{\partial V_{k}}=\sum_{i=k}^{N_{x}}(d_{i+1}-d_{i})\prod_{1\le t\le i, t\neq k}V_{t}.
\end{equation}
This guarantees back propagation of the projection layer. In order to speed up implementation, the processing of all rays are implemented in parallel via GPUs.

\noindent \textbf{Joint Training} Because our network consists of three subnetworks, we divide the entire training process into three stages to guarantee convergence: 1) The 3D convolution network is trained independently for scene completion; 2) With fixed parameters of the 3D convolution network, we train the 2D convolution network for depth image inpaintng under the guidance of 3D models; 3) We train the entire network jointly and fine tune it with all the parameters freed in 2D and 3D convolution networks.

The training data are generated based on the SUNCG synthetic scene dataset provided in ~\cite{song2017semantic}.
We first create $N$ depth images by rendering randomly selected scenes under randomly picked camera viewpoints. Each depth image $D$ is then converted to a point cloud $P$. Assuming $D$ is the projection of $P$ under the viewpoint $v$, we project $P$ to $m$ depth maps from $m$ randomly sampled views near $v$ to avoid causing large holes and to ensure that sufficient contextual information is available in the learning process.
Each training sample consists of a point cloud and one of its corresponding depth.

\subsection{Progressive Scene Completion}
\label{sec:dqn}
Given an incomplete point cloud $P_0$ that is converted from $D_0$ with respect to view $v_0$, we describe in this subsection how to obtain the optimal next view sequence $v_1, v_2,...,v_n$.
The problem is defined as a Markov decision process (MDP) consisting of state, action, reward, and an agent which takes actions during the process. The agent inputs the current state, outputs the corresponding optimal action, and receives the most reward from the environment. We train our agent using DQN \cite{mnih2015human}, an algorithm of deep reinforcement learning. The definition of the proposed MDP and the training procedure are given below.

\begin{figure}
	\centering
	\includegraphics[width=0.475\textwidth]{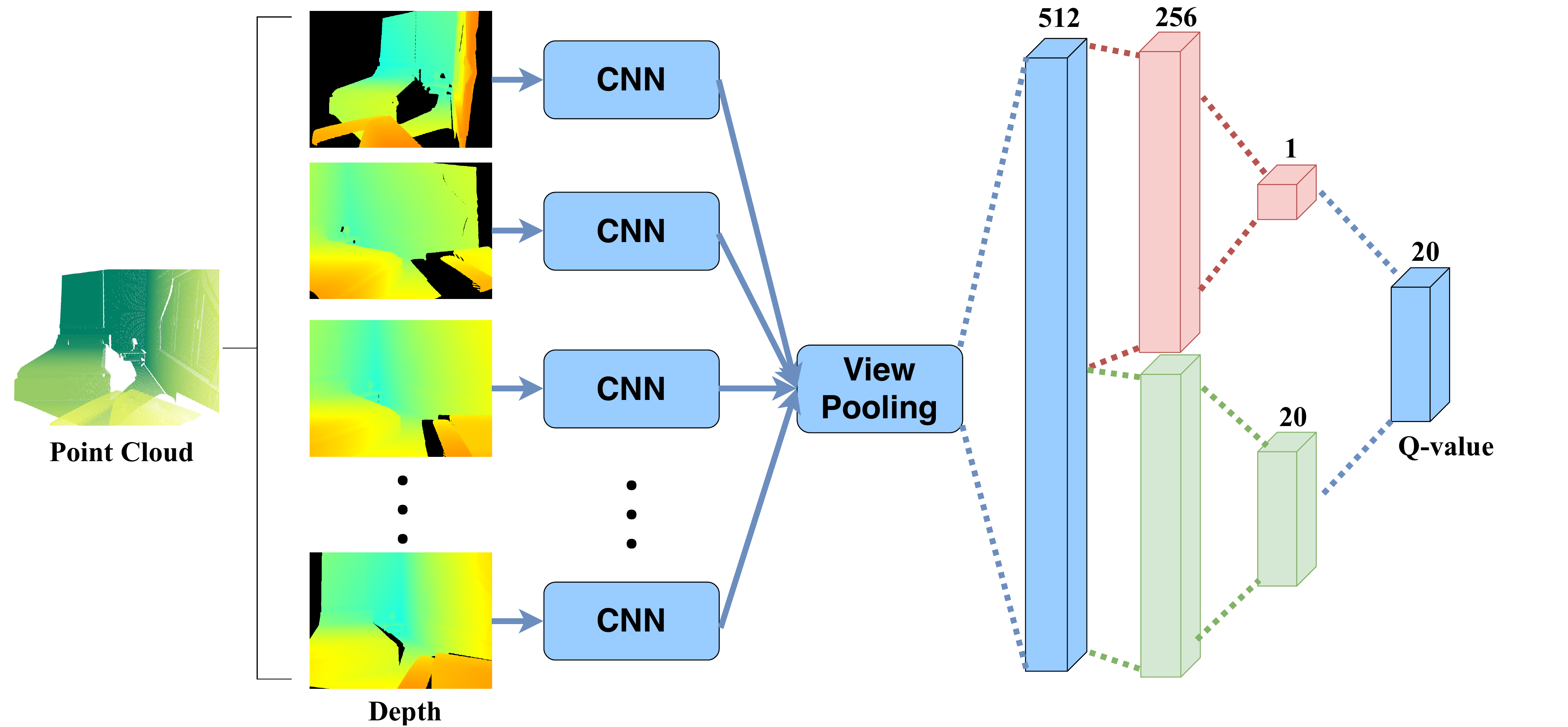}
	\caption{The architecture of our DQN. For a point cloud state, MVCNN is used to predict the best view for the next inpainting.}
	\label{fig:DQN}
\end{figure}

\noindent \textbf{State} We define the state as the updated point cloud at each iteration, with the initial state being $P_0$. As the iteration continues, the state for performing completion on the $i_{th}$ view is $P_{i-1}$, which is accumulated from all previous iteration updates.

\noindent \textbf{Action Space} The action at the $i_{th}$ iteration is to determine the next best view $v_i$. To ease the training process and support the use of DQN, we evenly sample a set of scene-centric camera views to form a discrete action space. Specifically, we first place $P_0$ in its bounding sphere and keep it upright. Then, two circle paths are created for both the equatorial and 45-degree latitude line. In our experiments, $20$ camera views are uniformly selected on these two paths, $10$ per circle. All views are facing to the center of the bounding sphere. We fixed these views for all training samples. The set of 20 views is denoted as $C = \{c_1,c_2,...,c_{20}\}$.

\noindent\textbf{Reward} An reward function is commonly unitized to evaluate the result for an action executed by the agent. In our work, at the $i_{th}$ iteration, the input is an incomplete depth map $D_i$ rendered from $P_{i-1}$ under view $v_i$ chosen in the action space $C$. The result of the agent action is an inpainted depth image $\hat{D_i}$. Hence the accuracy of this inpainting operation can be used as the primary rewarding strategy. It can be measured by the mean error of the pixels inside the holes between $\hat{D_i}$ and its ground truth $D^{gt}_i$. All the ground truth depth maps are pre-rendered from SUNCG dataset. Thus we define the award function as
\begin{equation}\label{equ1}
R_{i}^{acc}=-\frac{1}{|\Omega|}L^1_{\Omega}(\hat{D_i}, D^{gt}_i),
\end{equation}
where $L^1$ denotes the $ L_{1}$ loss, $\Omega$ the set of pixels inside the holes, and $|\Omega|$ the number of pixels inside $\Omega$.

\begin{figure}
	\centering
	\includegraphics[width=0.46\textwidth]{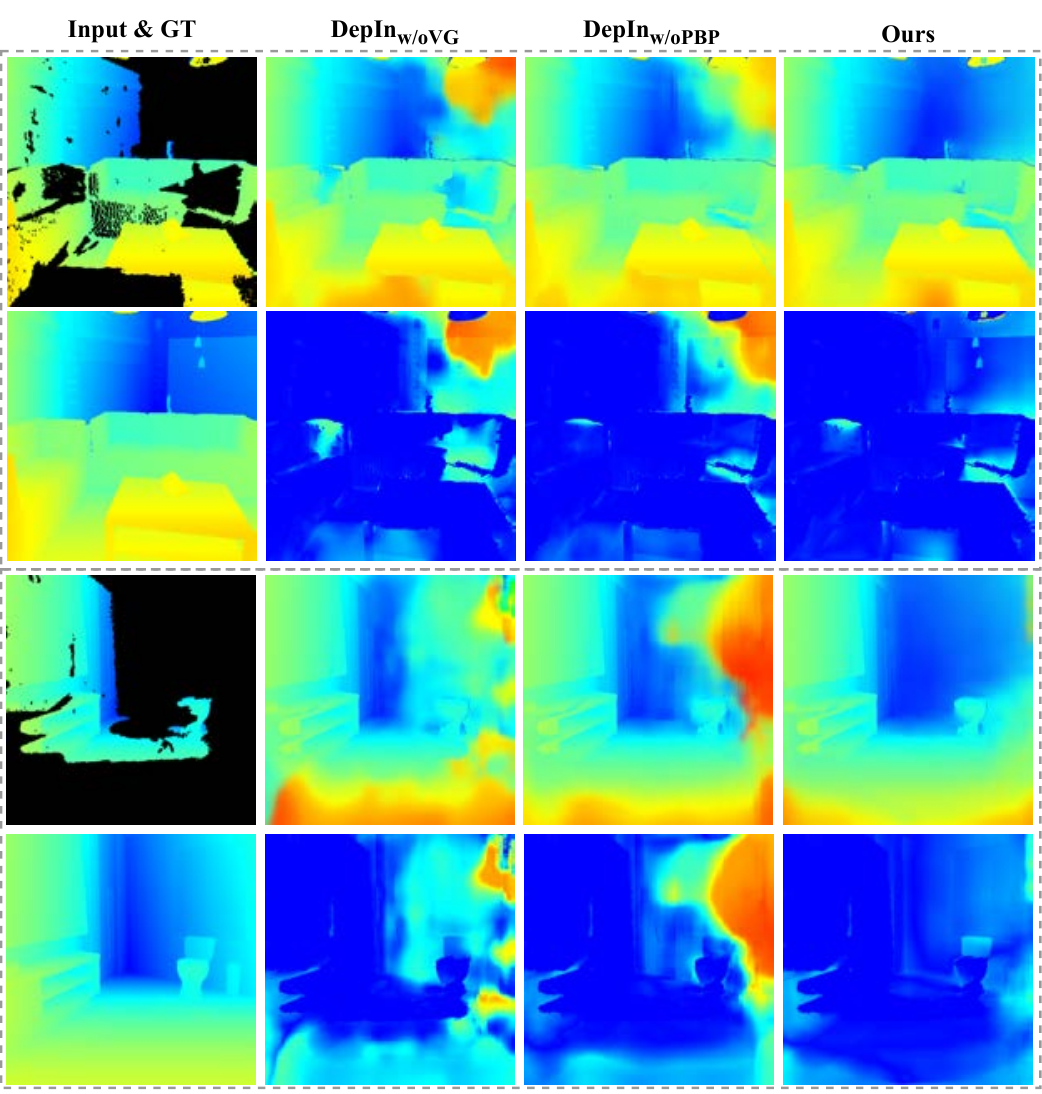}
	\caption{Comparisons on variants of depth inpainting network. Given incompleted depth images, we show results of our proposed method w/o volume-guidance, w/o projection back-propagation and also ours, compared with the groundtruth. Both the inpainted map and its error map are shown.}
	\label{fig:inpainting_ablation}
\end{figure}

If we only use the above reward function $R_{i}^{acc}$, the agent tends to change the viewpoint slightly in each action cycle, since doing this results in small holes. However, this incurs higher computational cost while accumulating errors. We thus introduce a new reward term to encourage inferring more missing points at each step.
This is implemented by measuring the percentage of filled original holes. To do so, we need to calculate the area of missing regions in an incomplete point cloud $P$, which is not trivial in a 3D space. Therefore, we project $P$ under all camera views to the action space $C$ and count the number of pixels inside the generated holes in each rendered image. The sum of these numbers is denoted as $Area^{h}(P)$ for measuring the area.
We thus define the new reward term as
\begin{equation}\label{equ2}
R_{i}^{hole}=\frac{Area^{h}(P_{i-1})-Area^{h}(P_{i})}{Area^{h}(P_0)}-1
\end{equation}
 to avoid the agent from choosing the same action as in previous steps.
 We further define a termination criterion to stop view path search by $ Area^{h}(P_{i})/Area^{h}(P_{0}) < 5\%$, which means that all missing points of $P_0$ have been nearly recovered. We set the reward for terminal to zero.

Therefore, our final reward function is
\begin{equation}\label{equ4}
R_{i}^{total}=wR_{i}^{acc}+(1-w)R_{i}^{hole},
\end{equation}
where $w$ is a fractional weight that balances the two reward terms.

\noindent \textbf{DQN Training} Our DQN is built upon MVCNN\cite{su15mvcnn}. It takes mutil-view depth maps projected from $ P_{i-1} $ as inputs and outputs the Q-value of different actions. The whole network is trained to approximate the action-value function $ Q(P_{i-1},v_{i}) $, which is the expected reward that the agent receives when taking action $ v_{i} $ at state $ P_{i-1} $.

To ensure stability of the learning process, we introduce a target network separated from the architecture of \cite{mnih2015human}, whose loss function for training DQN is
\begin{equation}\label{equ5}
Loss(\theta)=\mathbb{E}[(r+\gamma \max\limits_{v_{i+1}}Q(P_{i},v_{i+1};\theta')-Q(P_{i-1},v_{i};\theta))^{2}].
\end{equation}
where $ r $ is the reward, $ \gamma $ a discount factor, and $ \theta' $ the parameters of the target network. For effective learning, we create an experience replay buffer to reduce the correlation between data. The buffer stores the tuples $(P_{i-1},v_{i},r,P_{i}) $ proceeded with the episode. We also employ the technique of \cite{van2016deep} to remove upward bias caused by $ \max_{v_{i+1}}Q(P_{i},v_{i+1};\theta') $ and change the loss function to
\begin{equation}\label{equ6}
\begin{split}
\mathbb{L}_{our}&=\mathbb{E}[(r+\gamma Q(P_{i},\arg\max_{v_{i+1}}Q(P_{i},v_{i+1};\theta);\theta')\\
 &\quad-Q(P_{i-1},v_{i};\theta))^{2}].
\end{split}
\end{equation}
Combining with the dueling DQN structure~\cite{wang2015dueling}, our network structure is shown in Figure \ref{fig:DQN}.
At state $ P_{i-1} $, we render at all viewpoints $ c_{1}, c_{2}, ..., c_{20} $ in the action space $ C $ in $ 224\times224 $ resolution and get the corresponding multi-view depth maps $ D_{i}^{1}, D_{i}^{2}, ..., D_{i}^{20} $. These depth maps are then sent to the same $ CNN $ as inputs. After a view pooling layer and a fully-connected layer, we obtain a 512-D vector, which is split evenly into two parts to learn the advantage function $ A(v,P) $ and the state value function $ V(P) $~\cite{wang2015dueling}. Finally, after combining the results of the two functions, we have our final result, which is a 20-D Q-values based on the action space $ C $. We use an $\epsilon$-greedy policy to choose action $ v_{i} $ for state $ P_{i-1} $, i.e., a random action with probability $ 1-\epsilon $ or an action that maximizes the Q-values with probability $ \epsilon $. In the end, we reach the decision on depth map $ D_{i} $ for inpainting.

The training data are also generated from SUNCG. We use the same $ N $ depth images as in section \ref{sec:inpaint}. We also choose the action space $ C $ to generate new data. The ground truth depth maps, which are used in the reward calculation, are generated in the same viewpoint from the action space $ C $.

\begin{figure*}
	\centering
	\includegraphics[width=0.945\textwidth]{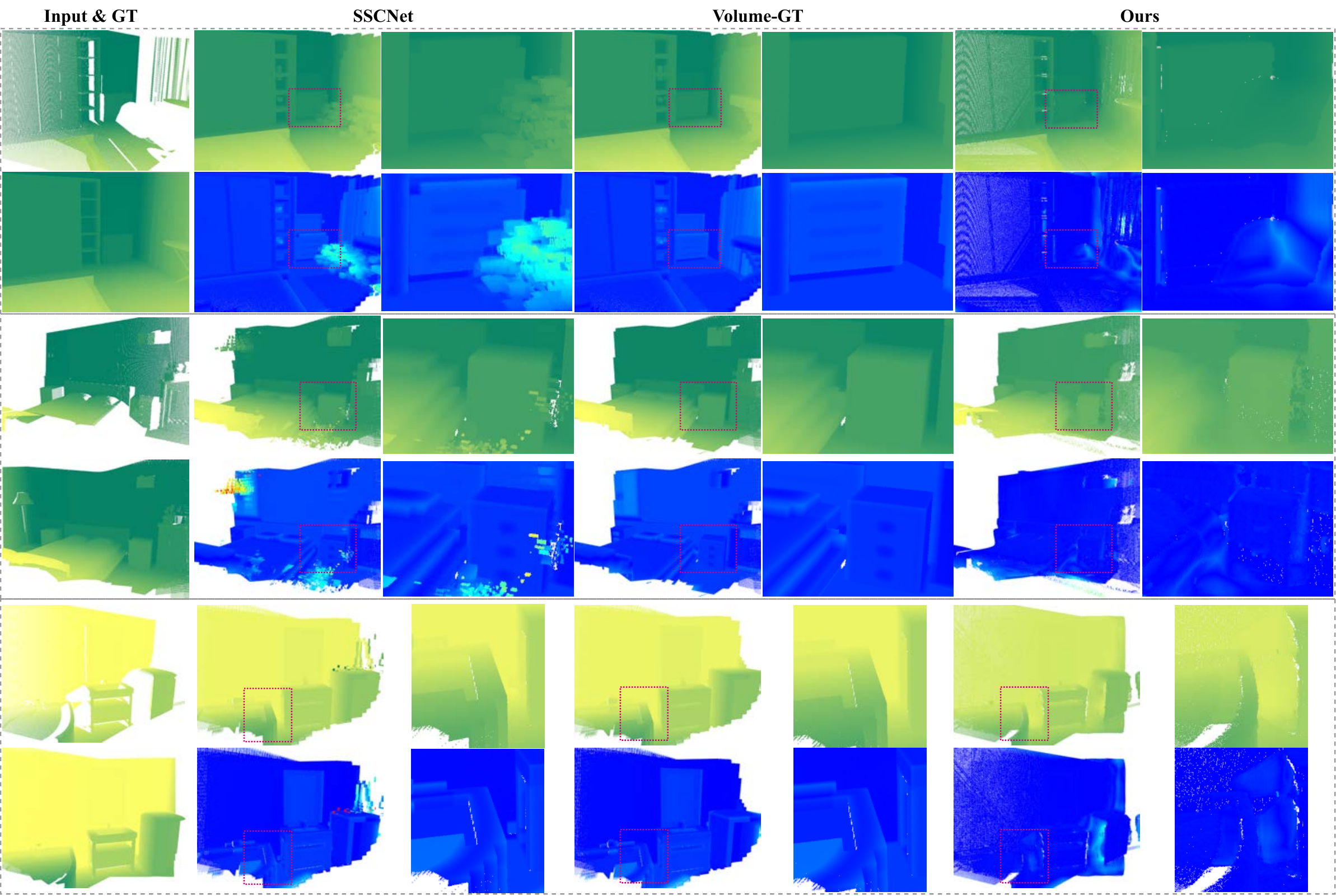}
	\caption{Comparisons against the state-of-the-arts. Given different inputs and the referenced groundtruth, we show the completion results of three methods, with the corresponding point cloud error maps below, and zoom-in areas beside.}
	\label{fig:method_comparison}
\end{figure*}

\begin{figure*}
	\centering
	\includegraphics[width=0.945\textwidth]{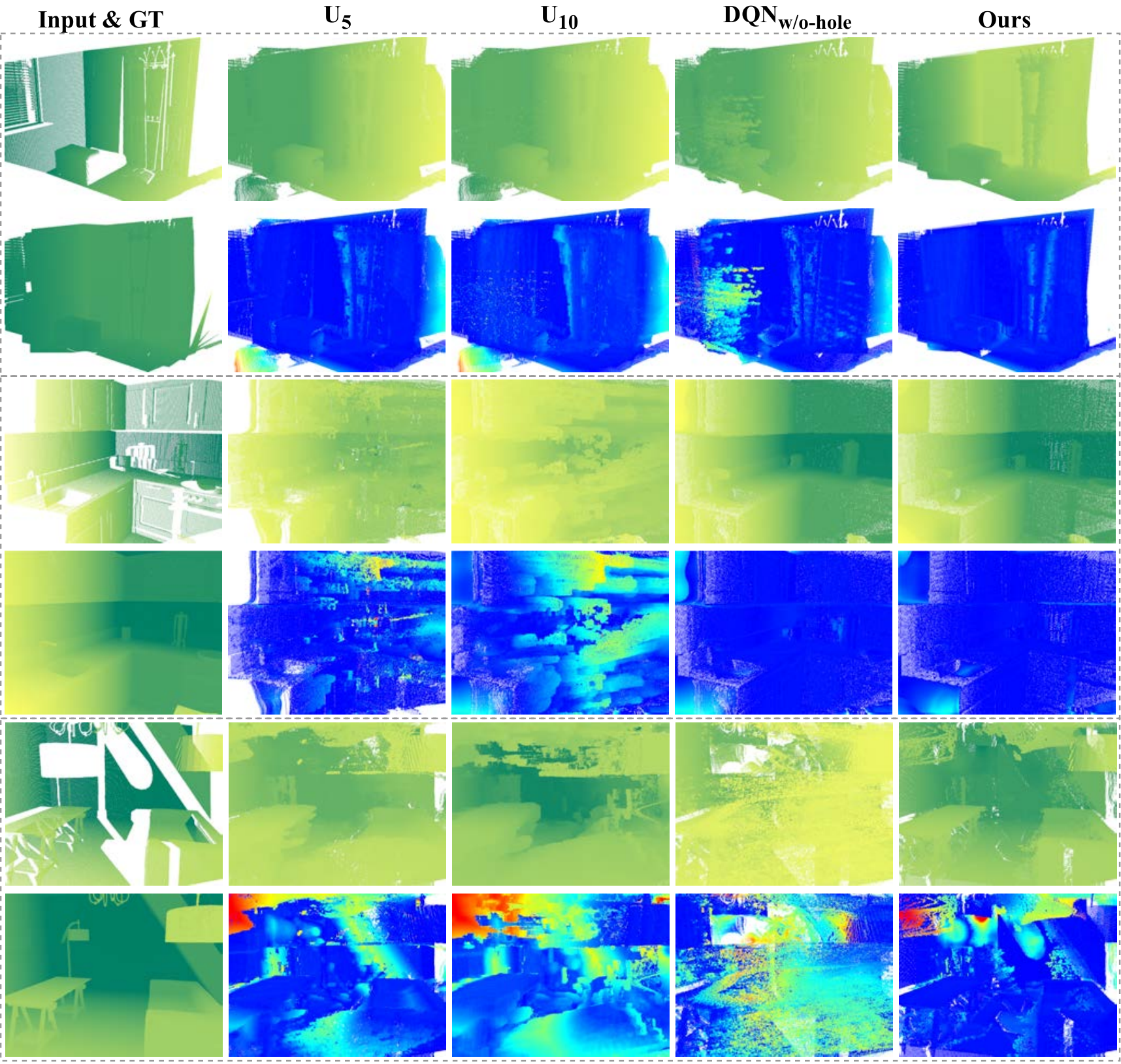}
	\caption{Comparisons on the variants of view path planning. Given different inputs and the referenced groundtruth, we show the completion results of four different approaches, with the corresponding point cloud error maps below.}
	\label{fig:DQN_comparison}
\end{figure*}

\section{Experimental Results}
\textbf{Dataset} The dataset we used to train our 2DCNN and DQN is generated from SUNCG \cite{song2017semantic}. Specifically, for 2DCNN, we set $N=3,000$ and $m=10$ and get $ 30,000$ depth maps. We further remove the maps whose camera views are occluded by doors or walls. Then, $3,000 $ of them are took for testing and the rest is used for training. For DQN, we set $ N = 2,500 $ with $ 2300 $ for the training episode and $ 200 $ for the testing.

\textbf{Implementation Details} Our network architecture is implemented in PyTorch. The provided pre-trained model of SSCNet \cite{song2017semantic} is used to initialize parameters of our 3DCNN part. It takes 30 hours to train inpainting network on our training dataset and 20 hours to fine-tune the whole network after the addition of projection layer. During DQN training process, we first use $ 200 $ episodes to fill experience replay buffer. In each episode, the DQN chooses the action randomly in each iteration step, and store the tuple $ (P_{i-1},v_{i},r,P_{i}) $ in the buffer. After those episodes being pre-trained, the network begins to learn by randomly sampled batches in buffers for each step during different episodes. The buffer can store $ 5,000 $ tuples and the batch size is set to 16. The weight $ w $ for reward calculation is set as $ 0.7 $ and the discount factor $ \gamma $ is set to $ 0.9 $, while $ \epsilon $ decreases from $ 0.9 $ to $ 0.2 $ over $ 10,000 $ steps and then be fixed to $ 0.2 $. Training DQN takes 3 days and running our complete algorithm once takes about $ 60s $ which adopts five view points on average.

\begin{table*}[htbp]	
	\centering
	\small
	\caption{Quantitative Comparisons against existing methods. The CD metric and the completeness metric (w.r.t different thresholds) are used.}
	\label{tab1}
	\begin{tabular}{ccccccccc}
		\hline
		& $SSCNet$ & $ Volume-GT_{1} $ & $ ScanComplete $ & $ Volume-GT_{2} $ & $ U_{5} $ & $ U_{10} $ & $DQN_{w/o-hole}$ & $ Ours $ \\
		\hline
		\hline		
		$ CD $ & 0.5162 & 0.5140 & 0.2193 & 0.2058 &  0.1642 & 0.1841 & 0.1495 &\textbf{0.1148} \\
		\hline
		$ C_{r=0.002}(\%) $ & 14.61 & 13.28 & 34.46 & 31.18 & 79.18 & 80.17 & 79.22 & 79.26 \\
		\hline
		$ C_{r=0.004}(\%) $ & 30.10 & 32.23 & 58.83& 61.11 & 83.33 & 84.15 & 83.50 & 83.68 \\
		\hline
		$ C_{r=0.006}(\%) $ & 52.82 & 50.14 & 74.60 & 74.88 & 85.81 & 86.56 & 86.02 & 86.28 \\
		\hline
		$ C_{r=0.008}(\%) $ & 71.24 & 72.33 & 79.59 & 81.04 & 87.66 & 88.33 & 87.81 & 88.20 \\
		\hline
		$ C_{r=0.010}(\%) $ & 78.23 & 78.96 & 81.01 & 81.61 & 89.06 & 89.70 & 89.24 & 89.68 \\
		\hline
	\end{tabular}
\end{table*}

\subsection{Comparisons Against State-of-the-Arts}
\label{sec:comparison}
In this part, we evaluate our proposed methods against SSCNet \cite{song2017semantic}, which is one of the most popular approaches in this area. Based on SSCNet, there although exists many incremental works such as ~\cite{wang2018adversarial} and ~\cite{guo2018view}, they all produce volumetric outputs in the same resolution as SSCNet. Regarding neither the code nor the pre-trained model of these methods is public, we propose to compare our result with the corresponding 3D groundtruth volume, whose output accuracy can be treated as the upper bound of all existing volume-based scene completion methods. We denote this method as volume-gt. For evaluation, we first render the volume obtained from SSCNet and volume-gt to several depth maps under the same viewpoints as our method. We then convert these depth maps to point cloud.

\textbf{Quantitative Comparisons} The Chamfer Distance (CD) \cite{fan2017point} is used as one of our metrics for evaluate the accuracy of our generated point set $P$, compared with the goundtruth point cloud $P_{GT}$. Similar to ~\cite{fan2017point}, we also use another completeness metric to evaluate how complete of the generated result. We define it as:
\begin{equation}\label{key}
C_{r}(P,P_{GT}) = \frac{\left| \{d(x,P)<r|x\in P_{GT}\}\right|}{\left| \{y | y\in P_{GT}\} \right|}
\end{equation}
where $ d(x,P) $ denotes the distance from a point $ x $ to a point set $ P $, $ \left|\cdot\right| $ denotes the number of the elements in the set, and $r$ means the distance threshold. In our experiments, we report the completeness w.r.t five different $r$ ($0.02,0.04,0.06,0.08,0.10$ are used). The results are reported in Tab \ref{tab1}. As seen, our approach significantly outperforms all the others. This also validates that the using of volumetric representation greatly reduces the quality of the outputs.

\textbf{Qualitative Comparisons} The visual comparisons of these methods are shown in Figure \ref{fig:method_comparison}. It can be seen that, the generated point cloud from SSCNet is of no surface details. Although our method shows more errors than volume-gt in some local regions, it overall produces more accurate results. This can be validated in Tab \ref{tab1}. In addition, by conducting completion in multiple views, our approach also recovers more missing points, showing better completeness as validated in Tab \ref{tab1}.
\subsection{Ablation Studies}
\label{sec:ablation}
To ensure the effectiveness of several key components of our system, we do some control experiments by removing each component.

\textbf{On Depth Inpainting} Firstly, to evaluate the efficacy of the volume guidance, we propose two variants of our method: 1) we train a 2D inpainting network directly without projecting volume as guidance, which is denoted as $DepIn_{w/oVG}$; 2) we train the volume guided 2D inpainting network without projection back-propagation, which is denoted as $DepIn_{w/oPBP}$. We use the metrics of $ L_{\Omega}^{1} $, $ PSNR $ and $ SSIM $ for the comparisons. The quantitative results are reported in Tab \ref{tab2} and the visual comparisons are shown in Figure \ref{fig:inpainting_ablation}. All of them show the superiority of our design.

\begin{table}[htbp]
	\centering
	\footnotesize
	\caption{Quantitative ablation studies on inpainting network.}
	\label{tab2}
	\begin{tabular}{cccc}
		\hline
		& $DepIn_{w/oVG}$ & $DepIn_{w/oPBP}$ & $ Ours $ \\
		\hline
		\hline
		$ L_{\Omega}^{1} $ & 0.0717 & 0.0574 & \textbf{0.0470}  \\
		
		$ PSNR $ & 22.15 & 23.12 & \textbf{24.73} \\
		
		$ SSIM $ & 0.910 & 0.926 & \textbf{0.930} \\
		\hline
	\end{tabular}
\end{table}

\textbf{On View Path Planning} Without using DQN for path planning, there exists a straightforward way to do completion: we can uniformly sample a fixed number of views from $C$ and directly perform depth implanting on them. In this uniform manner, two methods with two different numbers of views (5 and 10 are selected) are evaluated. We denote them as $U_5$ and $U_{10}$. The results of CD and $C_{r}(P, P_{GT})$ using these two methods and ours are reported in Tab \ref{tab1}. As seen, increasing the uniform sampled views causes accuracy reducing. This might be because of the increased accumulated errors. Using DQN greatly improves the accuracy, which validates the importance of a better view path. And all of them give rise to similar completeness. In addition, we also train a new DQN with only the reward $ R_{i}^{acc}$, denoted as $DQN_{w/o-hole}$, which chooses seven view points on average since it tends to pick views with small holes for higher $R_{i}^{acc}$. The results in Tab \ref{tab1} verify the efficiency of the reward $ R_{i}^{hole}$. Visual comparison results on some sampled scenes are shown in Figure ~\ref{fig:DQN_comparison}, where our proposed model results in much better appearances than others.

%

\section{Conclusion}
In this paper, we propose the first surface-based approach for 3D scene completion from a single depth image. The missing 3D points are inferred by conducting completion on multi-view depth maps. To guarantee a more accurate and consistent output, a volume-guided view inpianting network is proposed. In addition, a deep reinforcement learning framework is devised to seek the optimal view path to contribute the best result in accuracy. The experiments demonstrate that our model is the best choice and significantly outperforms existing methods. There are two research directions worth further exploration in the future: 1) how to make use of the texture information from the input RGBD images to achieve more accurate depth inpainting; 2) how to do texture completion together with the depth inpainting, to output a complete textured 3D scene.

\section*{Acknowledgements}
We thank the anonymous reviewers for the insightful and constructive comments.This work was funded in part by The Pearl River Talent Recruitment Program Innovative and Entrepreneurial Teams in 2017 under grant No. 2017ZT07X152, Shenzhen Fundamental Research Fund under grants No. KQTD2015033114415450 and No. ZDSYS201707251409055, and by the National Natural Science Foundation of China under Grant 91748104, Grant 61632006, Grant 61751203.

{\small
\bibliographystyle{ieee_fullname}
\bibliography{scecomp}
}

\clearpage
\section*{Supplemental Material}

\begin{figure*}[htbp]
	\centering
	\includegraphics[width=1\textwidth]{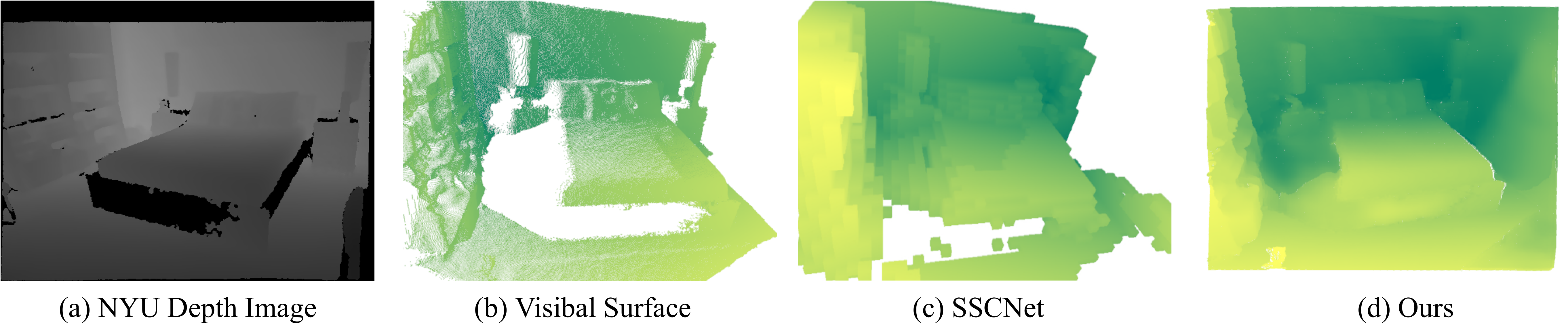}
	\caption{NYU data(a) testing results: SSCNet(c) and ours(d).}
	\label{fig:rebuttal}
\end{figure*}

\begin{figure*}[htbp]
	\centering
	\includegraphics[width=1\textwidth]{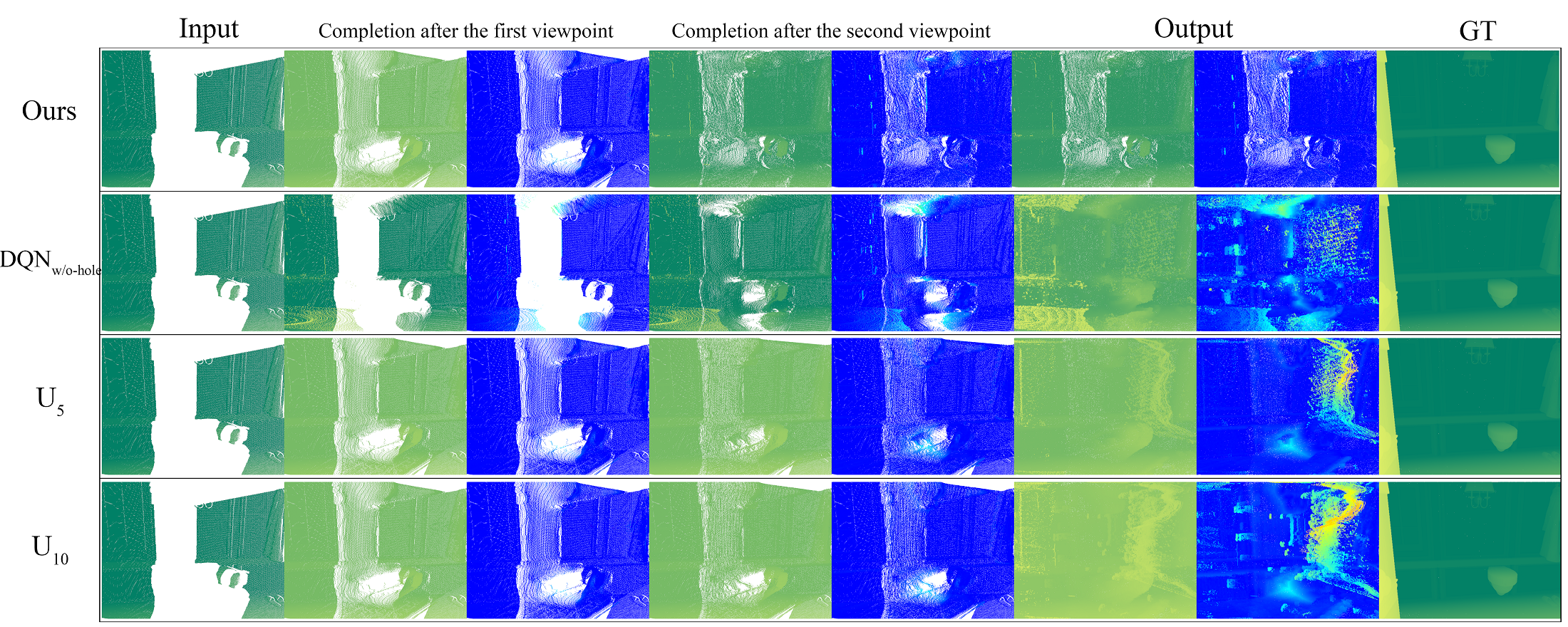}
	\caption{Comparisons of different methods choosing different view paths. Given the same input and the referenced groundtruth, we show the completion results after processing the first viewpoint and after the second viewpoint, and the final results where the whole view paths have been completed. The corresponding point cloud error maps are shown.}
	\label{fig:fig1}
\end{figure*}

\begin{figure*}[htbp]
	\centering
	\includegraphics[width=1\textwidth]{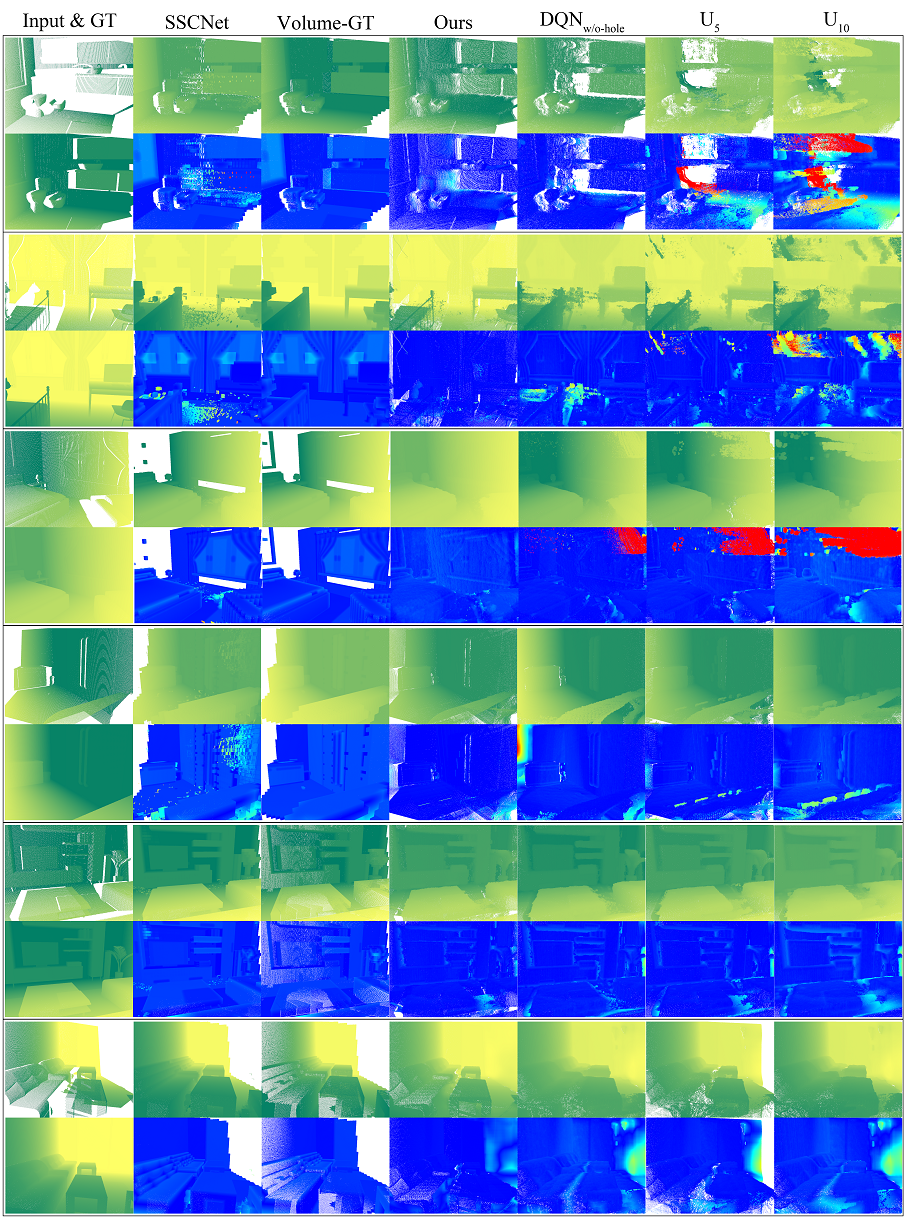}
	\caption{Comparisons of our method against the others. Given different inputs and the referenced groundtruth, we show the completion results of the six methods with the corresponding point cloud error maps shown below.}
	\label{fig:fig2}
\end{figure*}

\begin{figure*}[htbp]
	\centering
	\includegraphics[width=1\textwidth]{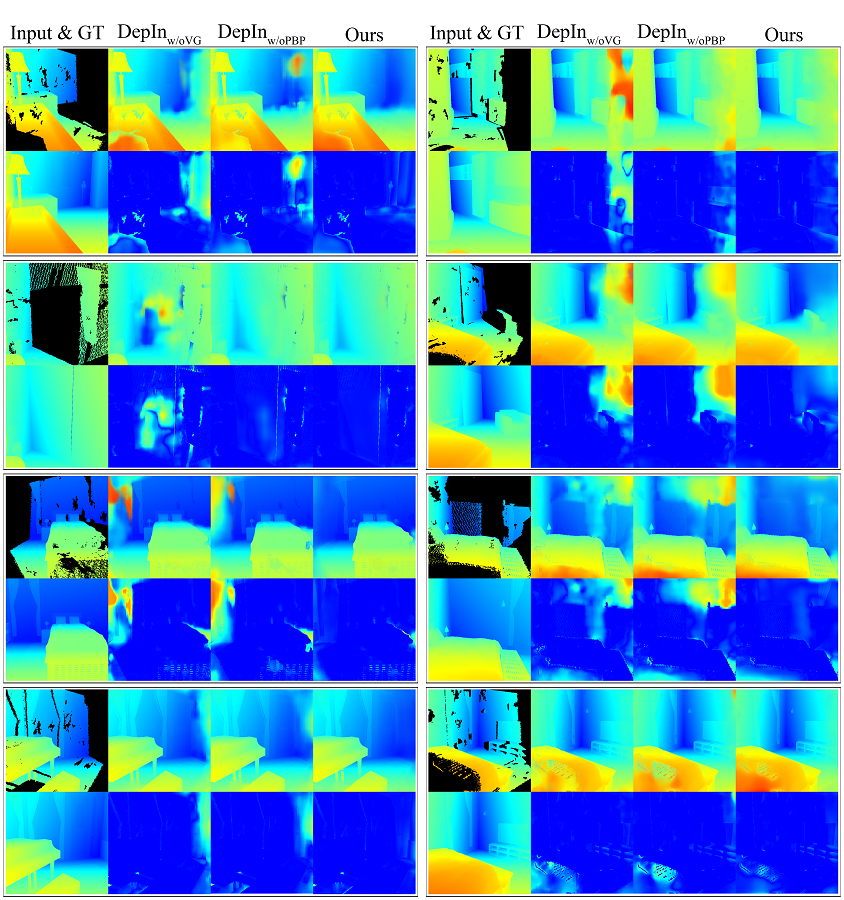}
	\caption{Comparisons on variants of depth inpainting network in eight groups.
		Given incompleted depth images, we show results of our proposed
		method with and without $1.)$ volume-guidance and $2.)$ projection back-propagation, compared with the groundtruth. The inpainted
		maps are shown in the first row and their error maps are shown below.}
	\label{fig:fig3}
\end{figure*}

In this supplemental material, more comparison results are shown: Fig \ref{fig:rebuttal} shows the results of our method and other methods testing on NYU dataset. Fig \ref{fig:fig1} shows comparisons of different methods selecting different view paths. Fig \ref{fig:fig2} shows more results, where our method is compared with voxel-based algorithms and other methods appearing in our paper. Fig \ref{fig:fig3} shows comparisons on all variants of our inpainting network. 

\end{document}